\documentclass{article}
\usepackage{colacl}

\usepackage{url}

\title{A Finite State and Data-Oriented Method for Grapheme to Phoneme Conversion}
\author{Gosse Bouma\\Alfa-informatica\\Rijksuniversiteit Groningen\\
Postbus 716\\
9700 AS Groningen\\
The Netherlands\\
{\it gosse@let.rug.nl}
}

\begin{document}

\maketitle
\begin{abstract}
  A finite-state method, based on leftmost longest-match replacement,
  is presented for segmenting words into graphemes, and for converting
  graphemes into phonemes. A small set of hand-crafted conversion
  rules for Dutch achieves a phoneme accuracy of over 93\%. The
  accuracy of the system is further improved by using
  transformation-based learning. The phoneme accuracy of the
  best system (using a large rule and a `lazy' variant of
  Brill's algoritm), trained on only 40K words, reaches 99\%.
\end{abstract}

\section{Introduction}

Automatic grapheme to phoneme conversion (i.e. the conversion of a
string of characters into a string of phonemes) is essential for
applications of text to speech synthesis dealing with unrestricted
text, where the input may contain words which do not occur in the system
dictionary. Furthermore, a transducer for grapheme to phoneme
conversion can be used to generate candidate replacements in a
(pronunciation-sensitive) spelling correction system. 
When given the pronunciation of a misspelled
word, the inverse of the grapheme to phoneme transducer will generate
all identically pronounced words.  Below, we present a method for
developing such grapheme to phoneme transducers based on a combination
of hand-crafted conversion rules, implemented using finite state
calculus, and automatically induced rules.

The hand-crafted system is defined as a two-step procedure:
segmentation of the input into a sequence of graphemes (i.e. sequences
of one or more characters typically corresponding to a single phoneme)
and conversion of graphemes into (sequences of) phonemes. 
The composition of the transducer which performs segmentation and the
transducer defined by the conversion rules, is a transducer which 
converts sequences of characters into sequences of phonemes.  

Specifying the conversion rules is a difficult task. Although
segmentation of the input can in principle be dispensed with, we found
that writing conversion rules for segmented input substantially
reduces the context-sensitivity and order-dependence of such rules.
We manually developed a grapheme to phoneme transducer for Dutch data 
obtained from {\sc celex} \cite{celex} and achieved a word
accuracy of 60.6\% and a phoneme accuracy of 93.6\%.

To improve the performance of our system, we used transformation-based
learning (TBL) \cite{Brill-CL}. Training data are obtained by aligning
the output of the hand-crafted finite state transducer with the
correct phoneme strings. These data can then be used as input for TBL,
provided that suitable rule templates are available. We performed
several experiments, in which the amount of training data, the
algorithm (Brill's original formulation and `lazy' variants
\cite{Samuel-coling}), and the number of rule templates varied. The
best experiment (40K words, using a `lazy' strategy with a large set
of rule templates) induces over 2000 transformation rules, leading to
92.6\% word accuracy and 99.0\% phoneme accuracy.  This result,
obtained using a relatively small set of training data, compares well
with that of other systems.

\section{Finite State Calculus}

\begin{figure*}[t]
\begin{center}
\begin{tabular}{ll}
{\tt []} & the empty string \\
{\tt [R$_1,\ldots ,R_n$]} & concatenation  \\
{\tt \{R$_1,\ldots ,R_n$\} } & disjunction \\
{\tt R\verb!^!} & optionality \\
{\tt ignore(A,B)} & ignore: A interspersed with elements of B \\
{\tt A x B}    & cross-product: the transducer which maps \\
               & all strings in A to all strings in B. \\
{\tt identity(A)} & identity: the transducer which maps each \\
                  & element in A onto itself. \\
{\tt T o U} & composition of the transducers T and U.\\
{\tt macro(Term,R)} & use Term as an abbreviation for R (where
                      Term and R may contain variables). 
\end{tabular}
\caption{A fragment of FSA regular expression syntax. 
A and B are regular expressions denoting recognizers,  T and U transducers, 
and R can be either.\label{fsa}}
\end{center}
\end{figure*}

As argued in \newcite{kaplan-kay}, \newcite{Karttunen95},
\newcite{Karttunen97}, and elsewhere, many of the rules used in
phonology and morphology can be analysed as special cases of regular
expressions. By extending the language of regular expressions with
operators which capture the interpretation of linguistic rule systems,
high-level linguistic descriptions can be compiled into finite state
automata directly. Furthermore, such automata can be combined with
other finite state automata performing low-level tasks such as
tokenization or lexical-lookup, or more advanced tasks such as shallow
parsing. Composition of the individual components into a single
transducer may lead to highly efficient processing.

The system described below was implemented using FSA Utilities,\footnote{
\url{www.let.rug.nl/~vannoord/fsa/}} 
a package for implementing and manipulating finite
state automata, which provides possibilities for defining new regular
expression operators. The part of FSA's built-in regular expression
syntax relevant to this paper, is listed in figure~\ref{fsa}.

One particular useful extension of the basic syntax of regular
expressions is the replace-operator. \newcite{Karttunen95} argues that
many phonological and morphological rules can be interpreted as rules
which replace a certain portion of the input string.  Although several
implementations of the replace-operator are proposed, the most
relevant case for our purposes is so-called 'leftmost
longest-match' replacement. In case of overlapping rule targets in the
input, this operator will replace the leftmost target, and in cases
where a rule target contains a prefix which is also a potential
target, the longer sequence will be replaced.
\newcite{GerdemannVanNoord} implement leftmost longest-match
replacement in FSA as the operator 

\begin{verbatim}
replace(Target, LeftContext,RightContext), 
\end{verbatim}

\noindent
where Target is a transducer defining the actual
replacement, and LeftContext and RightContext are regular expressions
defining the left- and rightcontext of the rule, respectively.

An example where leftmost replacement is useful is
hyphenation. Hyphenation of (non-compound) words in Dutch amounts to
segmenting a word into syllables, separated by hyphens. In cases where
(the written form of) a word can in principle be segmented in several
ways (i.e. the sequence 
{\tt alfabet} can be segmented as {\tt al-fa-bet, al-fab-et,
  alf-a-bet,} or {\tt  alf-ab-et}), the segmentation which maximizes
onsets is in general the correct one (i.e. {\tt al-fa-bet}). This
property of hyphenation is captured by leftmost replacement: 

\begin{verbatim}
macro(hyphenate, 
   replace([] x -, syllable, syllable)).
\end{verbatim}

\noindent
Leftmost replacement ensures that hyphens are introduced `eagerly', i.e.
as early as possible. Given a suitable definition of {\tt syllable},
this ensures that wherever a consonant can be final in a coda or
initial in the next onset, it is in fact added to the onset.

The segmentation task discussed below makes crucial use of longest match.

\section{A finite state method for grapheme to phoneme conversion}

Grapheme to phoneme conversion is implemented as the composition of
four transducers:

\begin{verbatim}
macro(graph2phon,     
   segmentation     % segment the input
 o mark_begin_end   % add '#'
 o conversion       % apply rules
 o clean_up      ). % remove markers         
\end{verbatim}

An example of conversion including the intermediate steps is given
below for the word {\tt aanknopingspunt} ({\em connection-point}).
 
\begin{quote}
\hspace{-25pt}
\begin{tabular}{ll}
{\bf\tt input}: &  {\tt aanknopingspunt} \\
{\bf\tt s}: &  {\tt aa-n-k-n-o-p-i-ng-s-p-u-n-t-} \\
{\bf\tt m}: &  {\tt \#-aa-n-k-n-o-p-i-ng-s-p-u-n-t-\#}\\
{\bf\tt co}: & {\tt \#-a+N+k-n-o-p-I+N+s-p-\}+n-t-\#}\\
{\bf\tt cl}: &  {\tt aNknopINsp\}nt}
\end{tabular}
\end{quote}

\noindent
The first transducer ({\tt {\bf s}egmentation}) takes as its input a sequence of
characters and groups these into segments. The second transducer 
({\tt{\bf m}ark\_begin\_end}) adds a
marker ('\#') to the beginning and end of the sequence of segments.
The third transducer ({\tt {\bf co}nversion}) performs the actual conversion step. It converts
each segment into a sequence of (zero or more) phonemes. The final
step ({\tt {\bf cl}ean\_up}) removes all markers. The output is a list of
phonemes in the notation used by {\sc celex} (which can be easily translated
into the more common {\sc sampa}-notation). 

\subsection{Segmentation}

The goal of segmentation is to divide a word into a sequence of
graphemes, providing a convenient input level of representation for the 
actual grapheme to phoneme conversion rules.

While there are many letter-combinations which are realized as a
single phoneme ({\tt ch, ng, aa, bb, ..}), it is only rarely the case
that a single letter is mapped onto more than one phoneme ({\tt x}),
or that a letter receives no pronunciation at all (such as word-final
{\tt n} in Dutch, which is elided if it is proceeded by a schwa). As
the number of cases where multiple letters have to be mapped onto a
single phoneme is relatively high, it is natural to model a letter to
phoneme system as involving two subtasks: segmentation and conversion.
Segmentation splits an input string into graphemes,
where each grapheme typically, but not necessarily, corresponds to a
single phoneme.

Segmentation is defined as:

\begin{verbatim}
macro(segmentation, 
 replace(
  [identity(graphemes), [] x - ],[],[]) 
         ).
\end{verbatim}

\noindent
The macro {\tt graphemes} defines the set of graphemes. It contains 
77 elements, some of which are:

\begin{verbatim}
a, aa, au, ai, aai, e, ee, ei, eu, eau, 
eeu, i, ie, iee, ieu, ij, o, oe, oei,..
\end{verbatim}

\noindent
Segmentation attaches the marker '{\tt -}' to each grapheme. 
Segmentation, as it is defined here, is not
context-sensitive, and thus the second and third arguments of replace
are simply empty.  As the set of graphemes contains many elements which
are substrings of other graphemes (i.e. {\tt e} is a substring of {\tt
  ei}, {\tt eau}, etc.), longest-match is essential: the segmentation
of {\tt beiaardier} ({\em carillon player}) should be {\tt
  b-ei-aa-r-d-ie-r-} and not {\tt b-e-i-a-a-r-d-i-e-r-}. This effect can
be obtained by making the segment itself part of the target of the replace
statement. Targets are identified using leftmost longest-match, and thus 
at each point in the input, only the longest valid segment is marked. 

The set of graphemes contains a number of elements which might seem
superfluous. The grapheme {\tt aai}, for instance, translates as {\tt
  aj}, a sequence which could also be derived on the basis of two
graphemes {\tt aa} and {\tt i}. However, if we leave out the segment
{\tt aai}, segmentation (using leftmost longest match) of words such as {\tt waaien}
({\em to blow}) would lead to the segmentation {\tt w-aa-ie-n}, which
is unnatural, as it would require an extra conversion rule for {\tt
  ie}. Using the grapheme {\tt aai} allows for two conversion rules which
always map {\tt aai} to {\tt aj} and  {\tt ie} goes to {\tt i}. 

Segmentation as defined above provides the intuitively correct result
in almost all cases, given a suitably defined set of graphemes. There
are some cases which are less natural, but which do not necessarily
lead to errors. The grapheme {\tt eu}, for instance, almost always
goes to {\tt `|'}, but translates as {\tt `e,j,\}'} in (loan-) words
such as {\tt museum} and {\tt petroleum}. One might argue that
a segmentation {\tt e-u-} is therefore required, but a special
conversion rule which covers these exceptional cases (i.e. {\tt eu}
followed by {\tt m}) can easily be formulated. Similarly, {\tt ng}
almost always translates as {\tt N}, but in some cases actually
represents the two graphemes {\tt n-g-}, as in {\tt
  aaneengesloten} ({\em connected}), where it should be translated as
{\tt NG}. This case is harder to detect, and is a potential source of errors. 

\subsection{The Conversion Rules}

The {\tt g2p} operator is designed to facilitate the 
formulation of conversion rules for segmented input: 

\begin{verbatim}
macro(g2p(Target,LtCont,RtCont),
  replace([Target, - x +], 
             [ignore(LtCont,{+,-}), {-,+}], 
                ignore(RtCont,{+,-}) 
          ) 
     ).
\end{verbatim}

\noindent
The g2p-operator implements a special purpose version of the 
replace-operator. The replacement of the marker '-' by '+' in 
the target ensures that g2p-conversion rules cannot
apply in sequence to the same grapheme.\footnote{Note that the input and output 
alphabet are not disjoint, and thus rules
applying in sequence to the same part of the input are not
excluded in principle.} Second, each target of the g2p-operator must be
a grapheme (and not some substring of it). This is a consequence of 
the fact that 
the final element of the left-context must be a marker and the target 
itself ends in '-'. 
Finally, the ignore statements in the left and right context imply
that the rule contexts can abstract over the potential presence of
markers. 

An overview of the conversion rules we used for Dutch is given in
Figure~\ref{conversion}. As the rules are applied in sequence,
exceptional rules can be ordered before the regular cases, thus
allowing the regular cases to be specified with little or no context.
The {\tt special\_vowel\_rules} deal with exceptional translations of
graphemes such as {\tt eu} or cases where {\tt i} or {\tt ij}
goes to '{\tt @}'. The {\tt short\_vowel\_rules} treat single vowels
preceding two consonants, or a word final consonant. One problematic
case is {\tt e}, which can be translated either as {\tt 'E'} or {\tt
  '@'}. Here, an approximation is attempted which specifies the
context where {\tt e} goes {\tt 'E'}, and subsumes the other case
under the general rule for short vowels. The {\tt
  special\_consonant\_rules} address devoicing and a few other
exceptional cases. The {\tt default\_rules} supply a default mapping
for a large number of graphemes. The target of this rule is a long
disjunction of grapheme-phoneme mappings. As this rule-set applies
after all more specific cases have been dealt with, no context
restrictions need to be specified.

\begin{figure*}
\begin{center}
\begin{verbatim}
macro(conversion,     special_vowel_rules o short_vowel_rules 
                    o special_consonant_rules o default_rules  ).
macro(special_vowel_rules,
    g2p([e,u] x [e,j,}], [],  m)  %% museum 
  o g2p(i     x @,       [],  g)  %% moedig(st)
  o g2p([i,j] x @,       l,   k)  %% mogelijkheid  
  .... ).
macro(short_vowel_rules,       
    g2p(e x 'E', [], {[t,t],[k,k],x,...})
    g2p({ a x 'A' , e x @, i x 'I' , o x 'O', u x '}' }, [], [cons, {cons , #}] ) 
     ).
macro(special_consonant_rules,
     g2p(b x p,        [], {s,t,#})  
   o g2p([d,t^] x t,   [], {s,g,k,j,v,h,z,#})        
   o g2p({ f x v, s x z},  [], {b,d})        
   o g2p(g x 'G',      vowel, vowel)         
   o g2p(n x 'N',      [], {k,q})
   o g2p(n x [],       [@],[#])
   ...).
macro(default_rules, 
  g2p({ [a,a] x a, [a,a,i] x [a,j], [a,u] x 'M',  [e,a,u] x o, ....,
         [b,b] x b, [d,d] x d, ...., [c,h] x 'x', [s,c,h] x [s,x], [n,g] x 'N', ... 
      }, [], [] )   
     ).
\end{verbatim}
\caption{Conversion Rules\label{conversion}}
\end{center}

\end{figure*}

Depending somewhat on how one counts, the full set of conversion rules
for Dutch contains approximately 80 conversion rules, more than 40 of
which are default mappings requiring no context.\footnote{It should be noted
that we only considered words which do not contain
diacritics. Including those is unproblematic in principle, but would
lead to a slight increase of the number of rules.} Compilation of the
complete system results in a (minimal, deterministic) transducer with
747 states and 20,123 transitions.

\subsection{Test results and discussion}

The accuracy of the hand-crafted system was evaluated by testing it on 
all of the words wihtout diacritics in the {\sc celex} lexical database which have a
phonetic transcription. After several
development cycles, we achieved a word accuracy of 60.6\% and a
phoneme accuracy (measured as the edit distance between the phoneme string
produced by the system and the correct string, divided by the number
of phonemes in the correct string) of 93.6\%. 

There have been relatively few attempts at developing grapheme to
phoneme conversion systems using finite state technology alone.
\newcite{williams} reports on a system for Welsh, which uses no less
than 700 rules implemented in a rather restricted environment. The
rules are also implemented in a two-level system, {\sc pc-kimmo},
\cite{pckimmo}, but this still requires over 400 rules.
\newcite{Mobius} report on full-fledged text-to-speech system for
German, containing around 200 rules (which are compiled into a
weighted finite state transducer) for the grapheme-to-phoneme
conversion step. These numbers suggest that our implementation (which
contains around 80 rules in total) benefits considerably from the
flexibility and high-level of abstraction made available by finite
state calculus.

One might suspect that a two-level 
approach to grapheme to phoneme conversion is more appropriate than
the sequential approach used here. Somewhat surprisingly, however,
Williams concludes that a sequential approach is
preferable. The formulation of rules in the latter approach is more
intuitive, and rule ordering provides a way of dealing with
exceptional cases which is not easily available in a two-level system. 

While further improvements would definitely have been possible at this
point, it becomes increasingly difficult to do this on the basis of
linguistic knowledge alone. That is, most of the rules which have to
be added deal with highly idiosyncratic cases (often related to
loan-words) which can only be discovered by browsing through the test
results of previous runs. At this point, switching from a
linguistics-oriented to a data-oriented methodology, seemed
appropriate.

\section{Transformation-based grapheme to phoneme conversion}
\label{tbl}

\newcite{Brill-CL} demonstrates that accurate part-of-speech tagging can
be learned by using a two-step process. First, a simple system is used
which assigns the most probable tag to each word. The results of the
system are aligned with the correct tags for some corpus of training
data. Next, (context-sensitive) transformation rules are selected from
a pool of rule patterns, which replace erroneous tags by correct tags.
The rule with the largest benefit on the training data (i.e. the rule
for which the number of corrections minus the number of newly
introduced mistakes, is the largest) is learned and applied to the
training data. This process continues until no more rules can be
found which lead to improvement (above a certain threshold).

Transformation-based learning (TBL) can be applied to the present problem as 
well.\footnote{\newcite{HosteColing} compare TBL to C5.0
\cite{Quinlan} on a similar task, i.e. the mapping of the
pronunciation of one regional variant of Dutch into another.}
In this case, the base-line system is the finite state
transducer described above, which can be used to produce a set of
phonemic transcriptions for a word list. Next, these results are
aligned with the correct transcriptions. In combination with 
suitable rule patterns, these data can be used as input for a 
TBL process. 
\subsection{Alignment}

TBL requires aligned data for training and testing. While alignment is
mostly trivial for part-of-speech tagging, this is not the case for
the present task. Aligning data for grapheme-to-phoneme conversion
amounts to aligning each part of the input (a sequence of characters)
with a part of the output (a sequence of phonemes).  As the length of
both sequences is not guaranteed to be equal, it must be possible to
align more than one character with a single phoneme (the usual case)
or a single character with more than one phoneme (the exceptional
case, i.e. 'x'). The alignment problem is often solved
\cite{Dutoit,daelemansBosch96} by allowing 'null' symbols in the
phoneme string, and introducing `compound' phonemes, such as 'ks' to
account for exceptional cases where a single character must be aligned
with two phonemes. 

As our finite state system already segments the input into graphemes, we have
adopted a strategy where {\em graphemes} instead of characters are aligned
with phoneme strings (see \newcite{LawrenceKaye} for a similar
approach). The correspondence between graphemes and phonemes
is usually one to one, but it is no problem to align a  grapheme with
two or more phonemes. Null symbols are only introduced in the output
if a grapheme, such as word-final {\tt 'n'}, is not realized
phonologically.

For TBL, the input actually has to be aligned both with the system
output as well as with the correct phoneme string. The first task can
be solved trivially: since our finite state system proceeds by first
segmenting the input into graphemes (sequences of characters), and
then transduces each grapheme into a sequence of phonemes, we can
obtain aligned data by simply aligning each grapheme with its
corresponding phoneme string. The input is segmented into
graphemes by doing the segmentation step of the finite state
transducer only.  The corresponding phoneme strings can be identified
by applying the conversion transducer to the segmented input, while keeping the
boundary symbols '-' and '+'. As a
consequence of the design of the conversion-rules, the resulting
sequence of separated phonemes sequences stands in a one-to-one
relationship to the graphemes. An example is shown in figure
\ref{alignment}, where {\sc gr} represents the grapheme segmented
string, and {\sc sp} the (system) phoneme strings produced by the
finite state transducer. Note that the final {\sc sp} cell contains
only a boundary marker, indicating that the grapheme '{\tt n}' is translated
into the null phoneme.

\begin{figure}
\begin{tabular}{|l|l@{\hspace{2pt}}l|l@{\hspace{2pt}}l|l@{\hspace{2pt}}l|l@{\hspace{2pt}}l|l@{\hspace{2pt}}l|l@{\hspace{2pt}}l|l@{\hspace{2pt}}l|l@{\hspace{2pt}}l|l@{\hspace{2pt}}l}
\hline
Word & \multicolumn{14}{l|}{aalbessen ({\em currants})}\\
\hline
{\sc gr}& aa&-&l&-&b&-&e&-&ss&-&e&-&n&- \\
\hline
{\sc sp} & a&+&l&-&b&-&@&+&s&+&@&+&&+\\
\hline
{\sc cp}&  a&&l&&b&&E&&s&&@&&&\\
\hline
\end{tabular}
\caption{Alignment}
\label{alignment}
\end{figure}

\begin{figure*}[t]
\begin{center}
\begin{tabular}{|ll|ll|rr|}
\hline 
method & training data & phoneme  & word     & induced  & CPU time \\
       & (words)       & accuracy & accuracy & rules    & (in minutes)
       \\
\hline
Base-line &            & 93.6     & 60.6 & & \\   
\hline
Brill     &  20K       & 98.0     & 86.1     & 447      & 162 \\ 
Brill     &  40K       & 98.4     & 88.9     & 812      & 858 \\
\hline
lazy(5)   &  20K       & 97.6     & 83.5     & 337      & 43 \\
lazy(5)   &  40K       & 98.2     & 87.0     & 701      & 190 \\ 
lazy(5)   &  60K       & 98.4     & 88.3     & 922      & 397 \\
\hline
lazy(10)  &  20K       & 97.7     & 84.3     & 368      & 83 \\
lazy(10)  &  40K       & 98.2     & 87.5     & 738      & 335 \\
lazy(10)  &  60K       & 98.4     & 88.9     & 974      & 711 \\
\hline
lazy(5)+  &  20K       & 98.6     & 89.8     & 1225     & 186 \\
lazy(5)+  &  40K       & 99.0     & 92.6     & 2221     & 603 \\
\hline 
\end{tabular}
\caption{Experimental Results using training data produced by graph2phon\label{stats}}
\end{center}
\end{figure*}

For the alignment between graphemes (and, idirectly, the system
output) and the correct phoneme strings (as found in Celex), we used
the 'hand-seeded' probabilistic alignment procedure described by
\newcite{Black98}. From the finite state conversion rules, a set of
possible {\em grapheme $\rightarrow$ phoneme sequence} mappings can be
derived. This {\em allowables}-set was extended with (exceptional)
mappings present in the correct data, but not in the hand-crafted
system. We computed all possible alignments between (segmented) words
and correct phoneme strings licenced by the {\em allowables}-set.
Next, probabilities for all allowed mappings were estimated on the
basis of all possible alignments, and the data was parsed again, now
picking the most probable alignment for each word.  To minimize the
number of words that could not be aligned, a maximum of one unseen
mapping (which was assigned a low probability) was allowed per word.
With this modification, only one out of 1000 words on average could
not be aligned.\footnote{Typical cases are loan words ({\em umpires})
  and letter words (i.e. abbreviations) ({\em abc}).} 
These words were discarded.The aligned phoneme string
for the example in figure~\ref{alignment} is shown in the bottom line.
Note that the final cell is empty, representing the null phoneme.

\subsection{The experiments}

For the experiments with TBL we used the $\mu$-{\sc tbl}-package
\cite{Lager}.  This Prolog implementation of TBL 
is considerably more efficient (up to ten times faster) than Brill's
original (C) implementation. The speed-up results mainly from
using Prolog's first-argument indexing to access large
quantities of data efficiently.

We constructed a set of 22 rule templates which replace a predicted 
phoneme with a (corrected) phoneme on the basis of the underlying segment,
and a context consisting either of phoneme strings, with a maximum
length of two on either side, or a context consisting of graphemes,
with a maximal length of 1 on either side. Using 
only 20K words (which corresponds to almost 180K segments), 
and Brill's algorithm, we achieved a phoneme accuracy 
of 98.0\% (see figure~\ref{stats}) on a test set of
20K words of unseen data.\footnote{
The statistics for less time consuming experiments were obtained by
10-fold cross-validation and for the more expensive experiments by
5-fold cross-validation.}
Going to 40K words resulted in 98.4\%
phoneme accuracy. Note, however, that in spite of the relative efficiency
of the implementation, CPU time also goes up sharply. 

\begin{figure*}
\begin{center}
\begin{tabular}{|ll|ll|rr|}
\hline 
method & training data & phoneme  & word     & induced  & CPU time \\
       & (words)       & accuracy & accuracy & rules    & (in minutes)
       \\
\hline
Base-line &            & 72.9     & 10.8 & & \\ 
\hline
lazy(5)   &  20K       & 97.3     & 81.6     &  691     & 133 \\
lazy(5)   &  40K       & 98.0     & 86.0     & 1075     & 705 \\
\hline
\end{tabular}
\caption{Experimental results using data based on frequency.\label{more-stats}}

\end{center}
\end{figure*}

The heavy computation costs of TBL are due to the fact that for each
error in the training data, all possible instantiations of the rule
templates which correct this error are generated, and for each of
these instantiated rules the score on the whole training set has to be
computed.  \newcite{Samuel-coling} therefore propose an efficient,
`lazy', alternative, based on Monte Carlo sampling of the rules. For
each error in the training set, only a sample of the rules is
considered which might correct it. As rules which correct a high
number of errors have a higher chance of being sampled at some point,
higher scoring rules are more likely to be generated than lower
scoring rules, but no exhaustive search is required. We experimented
with sampling sizes 5 and 10. As CPU requirements are more modest, we
managed to perform experiments on 60K words in this case, which lead
to results which are comparable with Brill's algoritm applied to 40K words.

Apart from being able to work with larger data sets, the `lazy'
strategy also has the advantage that it can cope with larger sets of
rule templates.  Brill's algorithm slows down quickly when the set of
rule templates is extended, but for an algorithm based on rule
sampling, this effect is much less severe. Thus, we also constructed a
set of 500 rule templates, containing transformation rules which
allowed up to three graphemes or phoneme sequences as left or right
context, and also allowed for disjunctive contexts (i.e. the context
must contain an {\tt 'a'} at the first or second position to the
right).  We used this rule set in combination with a `lazy' strategy
with sampling size 5 (lazy(5)+ in figure~\ref{stats}). This led to a
further improvement of phoneme accuracy to 99.0\%, and word accuracy
of 92.6\%, using only 40K words of training material.

Finally, we investigated what the contribution was of using a
relatively accurate training set. To this end, we constructed an
alternative training set, in which every segment was
associated with its most probable phoneme (where frequencies were
obtained from the aligned {\sc celex} data).  As shown in
figure~\ref{more-stats}, the initial accuracy for such as system is
much lower than that of the hand-crafted system.  The experimental
results, for the `lazy' algorithm with sampling size 5, show that the
phoneme accuracy for training on 20K words is 0.3\% less than for the
corresponding experiment in figure~\ref{stats}.  For 40K words, the
difference is still 0.2\%, which, in both cases, corresponds to a difference in error
rate of around 10\%. As might be expected, the number of induced
rules is much higher now, and thus {\sc cpu}-requirements also
increase substantially.

\section{Concluding remarks}

We have presented a method for grapheme to phoneme conversion, which
combines a hand-crafted finite state transducer with rules induced by
a transformation-based learning. An advantage of this method is that
it is able to achieve a high level of accuracy using relatively small
training sets.  \newcite{Busser}, for instance, uses a memory-based
learning strategy to achieve 90.1\% word accuracy on the same task,
but used 90\% of the {\sc celex} data (over 300K words) as training
set and a (character/phoneme) window size of 9. \newcite{Hoste00}
achieve a word accuracy of 95.7\% and a phoneme accuracy of 99.5\% on
the same task, using a combination of machine learning techniques, as
well as additional data obtained from a second dictionary.

Given the result of \newcite{Roche-Schabes-TBL}, an obvious next step
is to compile the induced rules into an actual transducer, and to
compose this with the hand-crafted transducer. It should be noted,
however, that the number of induced rules is quite large in some of the
experiments, so that the compilation procedure may require some
attention.

\end{document}